\documentclass[letterpaper]{article}
\usepackage{aaai2026} 
\usepackage{times} 
\usepackage{helvet} 
\usepackage{courier} 
\usepackage[hyphens]{url} 
\usepackage{graphicx} 
\usepackage{natbib} 
\usepackage{caption} 
\usepackage{amsmath}
\begin{document}

\title{
    \includegraphics[width=0.25\textwidth]{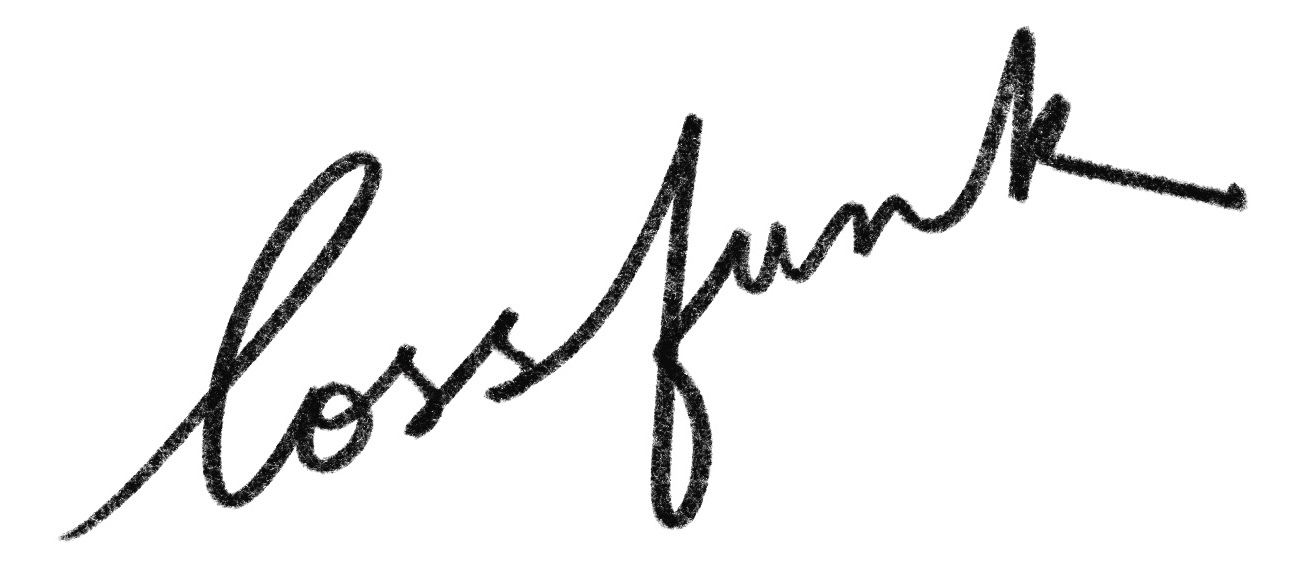}\\[0.75em]
    Future Is Unevenly Distributed\\[0.25em]
    {\large Forecasting Ability of LLMs Depends on What We’re Asking}
}

\author{
    Chinmay Karkar\textsuperscript{\rm 1},
    Paras Chopra\textsuperscript{\rm 1}
}
\affiliations{
    \vspace{0.75em}%
    \textsuperscript{\rm 1}Lossfunk\\[0.5em]
    chinmay.karkar@lossfunk.com, paras@lossfunk.com
}

\maketitle

\begin{abstract}
Large Language Models (LLMs) demonstrate partial forecasting competence across social, political, and economic events. Yet, their predictive ability varies sharply with domain structure and prompt framing. We investigate how forecasting performance varies with different model families on real-world questions about events that happened beyond the model cutoff date. We analyze how context, question type, and external knowledge affect accuracy and calibration, and how adding factual news context modifies belief formation and failure modes. Our results show that forecasting ability is highly variable as it depends on what, and how, we ask.
\end{abstract}

\section{Introduction}

Large Language Models (LLMs) have increasingly saturated a variety of benchmarks, demonstrating near-superhuman abilities in programming, mathematics, and scientific reasoning \cite{openai2025gpt5systemcard, deepseekr1, gemini25, claude45systemcard}. The forecasting abilities of LLMs, remain underexplored. With the growing integration of AI systems into high-stakes decision making, it is essential to assess whether such models can meaningfully predict real-world outcomes and to understand their systematic failure modes.
\begin{figure}[!t]
    \centering
    \includegraphics[width=\linewidth]{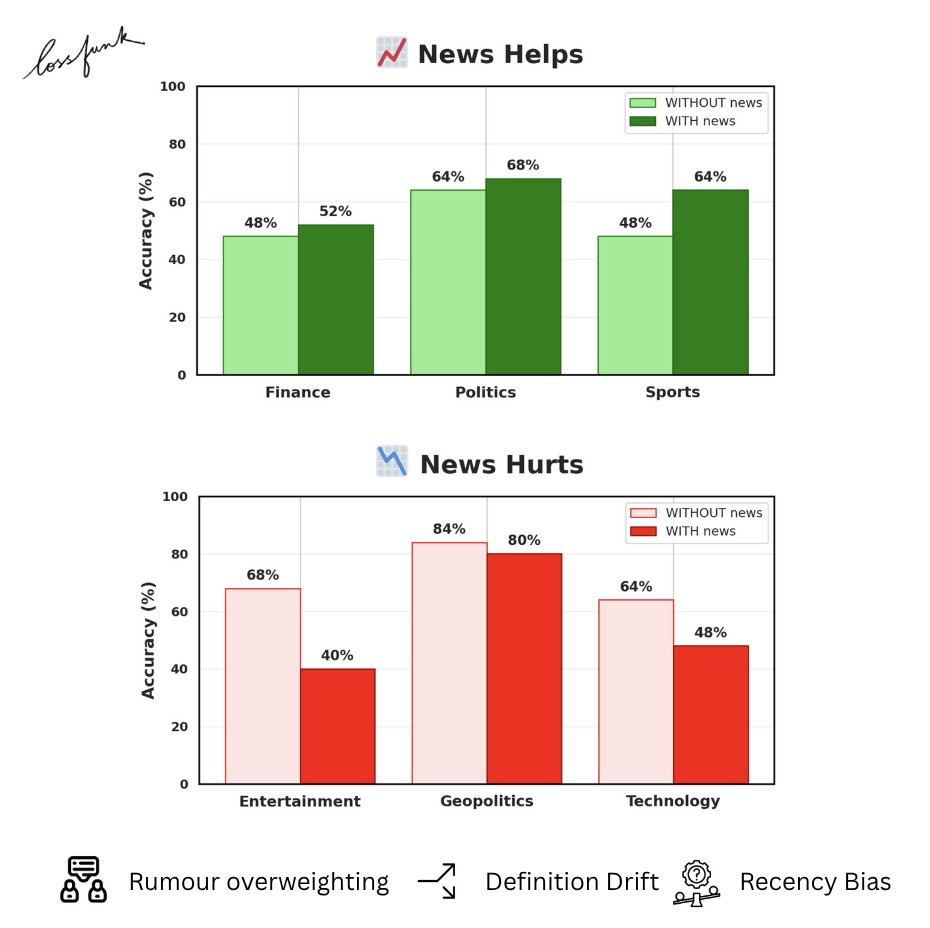}
    \caption{Impact of adding news context on category-wise forecasting accuracy (left: domains where news helps, right: domains where news hurts), along with a schematic summary of the main failure modes we observe: rumour overweighting, definition drift, and recency bias.}
    \label{fig:news_effects_overview}
\end{figure}
Recent investigations into LLM forecasting such as the Metaculus AI Forecasting Benchmark \cite{metaculusAIB} and ForecastBench \cite{karger2025forecastbenchdynamicbenchmarkai} examine whether models can perform in real-world prediction markets like Polymarket, Metaculus, and Kalshi \cite{polymarket,metaculus,kalshi}, and whether they can generate profit with external reasoning or retrieval tools. These studies, however, do not deeply analyze what types of questions LLMs answer confidently or where they fail. This gap in understanding limits our ability to interpret LLMs’ apparent forecasting success.

We address this gap by analyzing multiple LLM families, including both reasoning-optimized and non-reasoning models, and by evaluating them through a combination of qualitative and quantitative metrics such as accuracy and brier score. We also explore how model behavior changes when factual news context is added prior to prediction.

\subsection*{Contributions}
\begin{itemize}
    \item \textbf{Comprehensive evaluation:} We perform a qualitative and metric-based analysis of LLMs’ forecasting performance across multiple domains, both with and without contextual news inputs.
    \item \textbf{Failure mode taxonomy:} We identify and categorize recurrent failure modes that emerge during forecasting, particularly when contextual information is introduced, highlighting where reasoning and calibration diverge.
\end{itemize}


\section{Related Work}

The predictive reasoning capabilities of large language models have recently become a topic of growing interest. Early evidence from real-world forecasting tournaments showed that unassisted models such as GPT-4 underperformed relative to aggregate human forecasters \cite{schoenegger2023largelanguagemodelprediction}. Subsequent efforts have sought to improve this gap through large-scale fine-tuning and reinforcement learning on temporal reasoning tasks. Studies such as \cite{halawi2024approachinghumanlevelforecastinglanguage, lee2025advancingeventforecastingmassive, lu2025evaluatingllmsrealworldforecasting} demonstrate human-comparable accuracy, large-scale event forecasting training pipelines, and direct benchmarking of LLMs against expert forecasters, respectively. Collectively, these works indicate that iterative improvements in reasoning and retrieval alignment yield measurable forecasting gains.

Several recent initiatives have formalized AI forecasting evaluation through structured benchmarks. The \textit{Metaculus AI Forecast Benchmarking Tournament} \cite{metaculusAIB} and \textit{ForecastBench} \cite{karger2025forecastbenchdynamicbenchmarkai} present dynamic leaderboards using real prediction market questions drawn from platforms such as Polymarket and Metaculus \cite{polymarket,metaculus}.
 Prophet Arena \cite{yang2025llmasaprophetunderstandingpredictiveintelligence} further examines the theoretical grounding of “LLM-as-prophet” predictive intelligence, emphasizing calibration and model uncertainty. Alongside these developments, studies discuss key pitfalls in evaluating LLM forecasters, including logical leakage, unreliable news retrieval, and data contamination due to excessive reliance on model training cutoffs \cite{paleka2025pitfallsevaluatinglanguagemodel}.

Complementary datasets extend this line of inquiry toward temporal and contextual reasoning. \textit{ForecastQA} \cite{jin2021forecastqa}, \textit{Autocast} \cite{zou2022forecastingfutureworldevents}, \textit{ExpTime} \cite{yuan2024exptime}, \textit{FOReCAst} \cite{yuan2025forecastfutureoutcomereasoning}, and \textit{FutureX} \cite{zeng2025futurexadvancedlivebenchmark} each evaluate long-horizon prediction and confidence estimation under streaming updates. Mutschlechner and Jatowt (2025) analyze contextual cues in prompt design, finding that LLMs’ sensitivity to framing influences both calibration and directional correctness \cite{mutschlechner2025analyzingrolecontextforecasting}. 

Parallel to academic benchmarks, open-source infrastructures such as the \textit{Metaculus Forecasting Tools} \cite{metaculusTools2024} and \textit{ManifoldBot} \cite{micropredictionManifoldbot2024} enable autonomous LLM agents to interact directly with market-style systems, bridging probabilistic modeling with real-time trading and decision aggregation. Together, these works frame forecasting as an emerging dimension of LLM evaluation spanning Human–AI comparison, contextual robustness, and dynamic market participation.

\textbf{Our evaluation differs from these works as we study the failure modes of these models} with news added as context and also \textbf{show the clear difference in model performance} according to question category.

\section{Methodology}

In this section we detail the methodology behind our data processing and evaluation pipeline.

\subsection{Data Processing}

We began by collecting approximately 10{,}000 forecasting questions from various prediction markets such as Polymarket, Metaculus, and Manifold Markets \cite{polymarket,metaculus,manifold}, covering the period from January to July 2025. This period was chosen so that all questions selected were beyond the model's cutoff date.Many of these questions were noisy, that is, their context was hyper-localized or failed to test the forward-looking reasoning ability of large language models in a meaningful way. 

Some examples include:
\begin{quote}
\textit{“Daily coinflip”}\\[2pt]
\textit{“Will the \% chance of 'YES' on this market close above 50\%?”}\\
\textit{“Will I get a Donation/Payment of 10{,}000 or more Mana before 2025?”}
\end{quote}

\begin{figure}[htb]
    \centering
    \includegraphics[width=\linewidth]{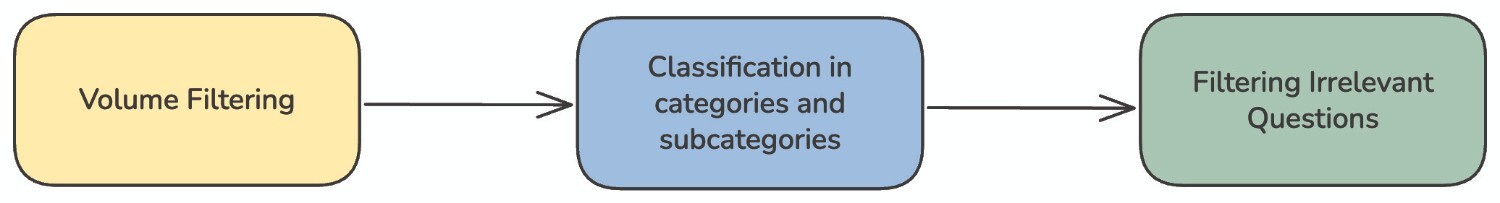}
    \caption{Overview of the data-processing pipeline used to construct the filtered forecasting benchmark.}
    \label{fig:pipeline}
\end{figure}

These questions do not provide any real signal of forecasting competence or reveal systematic failure modes. To extract a meaningful subset, we designed a three-stage filtering and classification pipeline (Figure~\ref{fig:pipeline}).

First, we applied \textbf{volume filtering} to remove low-liquidity markets, which typically correspond to hyper-personalized or creator-specific questions. Next, we employed \textit{Gemini 2.5 Flash} \cite{gemini25} as an LLM-as-a-Judge \cite{zheng2023judgingllmasajudgemtbenchchatbot} with the following prompt~(see Appendix~\ref{app:classifier}) to classify each question into six primary categories, each with five sub-categories:

\begin{itemize}
    \item \textbf{Politics}: Domestic Policy, Elections \& Campaigns, Political Parties \& Ideologies, Government Structure, Public Policy \& Social Issues
    \item \textbf{Entertainment}: Movies \& Television, Music \& Audio, Gaming, Celebrity \& Pop Culture, Books \& Literature
    \item \textbf{Sports}: Professional Sports, International Competitions, Individual Sports, Team Sports, Sports Culture \& Recreation
    \item \textbf{Technology}: Computing \& Software, Internet \& Digital Services, Mobile \& Consumer Electronics, Emerging Technologies, Tech Industry \& Business
    \item \textbf{Finance}: Personal Finance, Banking \& Financial Services, Markets \& Trading, Economic Indicators, Corporate Finance
    \item \textbf{Geopolitics}: International Relations, Global Conflicts, Trade \& Economics, Regional Affairs, Global Governance
\end{itemize}

Questions that did not align with any of the above were tagged as \textit{irrelevant}, reducing the corpus to roughly 700 items after aggressive filtering.  
Despite this reduction, certain residual questions remained non-event-based and failed to meaningfully test predictive reasoning. For instance:
\begin{quote}
\textit{“Will @Soaffine be active on Manifold again before April?”}
\end{quote}

To address these, we performed a second LLM-based filtering pass using a refined judging prompt~(see Appendix~\ref{app:filter}) to exclude localized or non-forecasting items. The final curated dataset contained 392 questions, evenly distributed across the categories and sub-categories listed above. For each retained question, we also preserved metadata such as \texttt{creationTime}, \texttt{resolutionTime}, and final resolution probability.
\subsection{Evaluation Methodology}

We start by sampling a uniform subset of 150 questions with seed 42 from the final corpus, ensuring an equal number of questions per category to construct a balanced evaluation dataset. This subset enables consistent cross-category comparison while maintaining representativeness of the larger filtered corpus.

We evaluate a mixture of reasoning-focused and non-reasoning large language models: {GPT-5} \cite{openai2025gpt5systemcard}, {GPT-4.1} \cite{openai2024gpt41systemcard}, {DeepSeek-R1} \cite{deepseekr1}, and {Claude 3.7 Sonnet} \cite{claude37sonnetsystemcard}. We sample from the models at a temperature of 0.0, max token budget as 4500 tokens to ensure that models have enough buffer to express their reasoning, as well as for deterministic sampling with 0.0 temperature. 
  
Each model is prompted using a standard forecasting prompt ~(see Appendix~\ref{app:coreprompt}) along with the question text and its creation date, to provide temporal grounding. Apart from this contextual timestamp, the models have no access to external tools, retrieval systems, or web search capabilities.

For every prompt, each LLM outputs two required fields:

\begin{quote}
\texttt{<answer>}YES/NO\texttt{</answer>}\\
\texttt{<conf>}0--1 confidence score\texttt{</conf>}
\end{quote}

We evaluate predictions using three key metrics: \textbf{accuracy}, the \textbf{brier score}, and the \textbf{Expected Calibration Error (ECE)}.

\textbf{Accuracy.} Measures whether the model’s predicted resolution (\texttt{<answer>}) matches the actual market resolution for each question. A correct match contributes 1, and an incorrect match contributes 0; the mean across all samples yields the final accuracy score.

\textbf{Brier Score.} Quantifies probabilistic calibration by penalizing confidence errors and is formally defined as:
\begin{equation}
\text{Brier Score} = \frac{1}{N}\sum_{i=1}^{N}(f_i - o_i)^2,
\end{equation}
where \( f_i \) denotes the model’s predicted probability (confidence) for a “YES” outcome, and \( o_i \in \{0,1\} \) represents the ground-truth outcome. Lower values indicate better probabilistic accuracy.

\textbf{Expected Calibration Error (ECE).} Measures the discrepancy between predicted confidence and empirical accuracy across probability bins. Predictions are partitioned into \( M \) bins based on confidence, and ECE is computed as:
\begin{equation}
\text{ECE} = \sum_{m=1}^{M} \frac{|B_m|}{N}\, \big|\text{acc}(B_m) - \text{conf}(B_m)\big|,
\end{equation}
where \( B_m \) is the set of predictions whose confidence scores fall into bin \( m \), \( \text{acc}(B_m) \) is the average accuracy within that bin, and \( \text{conf}(B_m) \) is the mean predicted confidence. Lower ECE values indicate better calibration between confidence and correctness.

\subsection{Evaluation with News Context}

For the second evaluation condition, we augment each forecasting question with external context retrieved from contemporary news sources.This ensures that LLMs have the same set of information that a human forecaster would have, when the event was created as a question in the prediction market. We collect ten news snippets per question by querying \textit{Exa} \cite{exa2025api} with the question text, using the question’s creation date as the upper bound for publication time. Despite the temporal cutoff, we occasionally observed leakage in the form of articles published after the creation date. Such snippets were removed completely to ensure temporal purity in all model inputs.

Each model is then re-evaluated on the context-augmented prompt~(see Appendix~\ref{app:newsprompt}) using the same scoring metrics, \textit{Accuracy}, \textit{Brier score} and \textit{ECE}, to measure how additional factual context influences forecasting calibration and directional correctness.

\section{Results by Category}

Table~\ref{tab:category_metrics_wo_news} presents category-wise model performance \textit{without news context}, evaluated using Accuracy, Brier Score, and Expected Calibration Error (ECE).  
ECE captures the deviation between model-predicted probabilities and observed outcomes, offering a finer measure of calibration quality beyond raw accuracy.  
Results show that GPT-5 and Claude-3.7 achieve strong calibration on structured domains such as \textit{Geopolitics} and \textit{Politics}, while DeepSeek-R1, GPT-4.1 display higher ECE in noisier domains like \textit{Entertainment} and \textit{Technology}.

\begin{table}[!t]
\centering
{\small
\setlength{\tabcolsep}{1mm}
\renewcommand{\arraystretch}{1.05}
\begin{tabular}{llccc}
\hline
\textbf{Category} & \textbf{Model} & \textbf{Accuracy} & \textbf{Brier} & \textbf{ECE} \\
\hline
Entertainment & Claude-3.7 & 68.00\% & 0.23 & 0.18 \\
              & DeepSeek-R1 & 68.00\% & 0.28 & 0.20 \\
              & GPT-4.1     & 64.00\% & 0.33 & 0.26 \\
              & GPT-5       & 52.00\% & 0.24 & 0.20 \\
\hline
Finance       & Claude-3.7 & 44.00\% & 0.31 & 0.31 \\
              & DeepSeek-R1 & 48.00\% & 0.35 & 0.29 \\
              & GPT-4.1     & 40.00\% & 0.33 & 0.27 \\
              & GPT-5       & 56.00\% & 0.26 & 0.20 \\
\hline
Geopolitics   & Claude-3.7 & 84.00\% & 0.12 & 0.12 \\
              & DeepSeek-R1 & 84.00\% & 0.32 & 0.36 \\
              & GPT-4.1     & 88.00\% & 0.40 & 0.45 \\
              & GPT-5       & 84.00\% & 0.14 & 0.09 \\
\hline
Politics      & Claude-3.7 & 68.00\% & 0.22 & 0.25 \\
              & DeepSeek-R1 & 64.00\% & 0.27 & 0.29 \\
              & GPT-4.1     & 72.00\% & 0.38 & 0.42 \\
              & GPT-5       & 64.00\% & 0.21 & 0.17 \\
\hline
Sports        & Claude-3.7 & 48.00\% & 0.28 & 0.33 \\
              & DeepSeek-R1 & 48.00\% & 0.26 & 0.23 \\
              & GPT-4.1     & 60.00\% & 0.45 & 0.49 \\
              & GPT-5       & 52.00\% & 0.28 & 0.26 \\
\hline
Technology    & Claude-3.7 & 68.00\% & 0.25 & 0.27 \\
              & DeepSeek-R1 & 64.00\% & 0.27 & 0.35 \\
              & GPT-4.1     & 72.00\% & 0.42 & 0.47 \\
              & GPT-5       & 68.00\% & 0.24 & 0.23 \\
\hline
\end{tabular}
}
\caption{Category-wise metrics \textit{without news context}. Each category contains 25 questions. Accuracy (\%) is shown alongside Brier and Expected Calibration Error (ECE) averaged per category.}
\label{tab:category_metrics_wo_news}
\end{table}

\section{News-Augmented Forecasting}

We next evaluate the same set of models when each question is supplemented with up to ten time-bounded news snippets retrieved prior to the question’s creation date.  
Table~\ref{tab:category_metrics_w_news} shows the corresponding metrics.  
While certain domains such as \textit{Finance} and \textit{Sports} benefit from context (lower brier, improved ECE), others such as \textit{Entertainment} and \textit{Technology} show declines, consistent with recency bias and noise amplification effects introduced by the additional text. We detail the explanation of these results in further sections. 

\begin{table}[!t]
\centering
{\small
\setlength{\tabcolsep}{1mm}
\renewcommand{\arraystretch}{1.05}
\begin{tabular}{llccc}
\hline
\textbf{Category} & \textbf{Model} & \textbf{Accuracy} & \textbf{Brier} & \textbf{ECE} \\
\hline
Entertainment & Claude-3.7 & 56.00\% & 0.27 & 0.19 \\
              & DeepSeek-R1 & 40.00\% & 0.34 & 0.42 \\
              & GPT-4.1     & 44.00\% & 0.36 & 0.35 \\
              & GPT-5       & 56.00\% & 0.27 & 0.25 \\
\hline
Finance       & Claude-3.7 & 56.00\% & 0.31 & 0.30 \\
              & DeepSeek-R1 & 52.00\% & 0.31 & 0.27 \\
              & GPT-4.1     & 68.00\% & 0.29 & 0.23 \\
              & GPT-5       & 60.00\% & 0.23 & 0.14 \\
\hline
Geopolitics   & Claude-3.7 & 80.00\% & 0.15 & 0.21 \\
              & DeepSeek-R1 & 80.00\% & 0.31 & 0.40 \\
              & GPT-4.1     & 76.00\% & 0.46 & 0.54 \\
              & GPT-5       & 84.00\% & 0.13 & 0.16 \\
\hline
Politics      & Claude-3.7 & 64.00\% & 0.26 & 0.26 \\
              & DeepSeek-R1 & 68.00\% & 0.29 & 0.33 \\
              & GPT-4.1     & 68.00\% & 0.33 & 0.31 \\
              & GPT-5       & 72.00\% & 0.18 & 0.12 \\
\hline
Sports        & Claude-3.7 & 60.00\% & 0.22 & 0.19 \\
              & DeepSeek-R1 & 64.00\% & 0.24 & 0.31 \\
              & GPT-4.1     & 56.00\% & 0.27 & 0.26 \\
              & GPT-5       & 56.00\% & 0.23 & 0.25 \\
\hline
Technology    & Claude-3.7 & 52.00\% & 0.29 & 0.36 \\
              & DeepSeek-R1 & 48.00\% & 0.33 & 0.40 \\
              & GPT-4.1     & 64.00\% & 0.43 & 0.53 \\
              & GPT-5       & 68.00\% & 0.23 & 0.30 \\
\hline
\end{tabular}
}
\caption{Category-wise metrics \textit{with news context}. Each category contains 25 questions. Accuracy (\%) is shown alongside Brier and Expected Calibration Error (ECE) averaged per category.}
\label{tab:category_metrics_w_news}
\end{table}

\section{Analysis and Failure Modes with Context}

Analyzing the models’ responses and reasoning traces from our evaluation reveals several recurring failure modes. When incorporating news as additional context, we observe issues consistent with those reported by Paleka et al. (2025) \cite{paleka2025pitfallsevaluatinglanguagemodel}, particularly those concerning unreliable news retrieval. Despite enforcing explicit temporal bounds on article publication dates through \textit{Exa} \cite{exa2025api}, we find that articles published after the question’s cutoff sometimes containing information that effectively resolves the question can still appear in the retrieved set when filtering is insufficient.

The addition of news as context improves the model in certain aspects such as clarifying the time scope of the question and latching onto proper signals, but it is also highly prone to various issues. We detail some of them below.

\textbf{Recency Bias.}
Models tend to overweight recent news over historical trends encoded during pretraining. This often leads to situations where the model changes a correct resolution into an incorrect one.

\begin{quote}
\textit{Question: “S\&P 500 above 6050 on June 13?”}\\[2pt]
\textbf{Raw model (a):} NO, 0.34 confidence, reasons that the index is near resistance at 6000 and mean reversion plus limited trading days make a breakout unlikely. (Correct)\\[3pt]
\textbf{News model (b):} YES, 0.54 confidence, reads snippets from the few days before June 13 describing the S\&P “flirting with 6000,” “record highs,” and “strategist upgrades targeting 6100.” (Wrong)
\end{quote}

The model allowed the most recent headlines to dominate its prior, turning a correct mean-reversion call into an overconfident breakout bet. The complete reasoning trace is provided~(see Appendix~\ref{app:recency}).

\textbf{Rumour Overweighting.}
Models frequently anchor to unverified information or speculation present in retrieved snippets, causing them to switch from a correct to an incorrect resolution.

\begin{quote}
\textit{Question: “Tariffs on China above 150\% by end of June?”}\\[2pt]
\textbf{Raw model (a):} NO, high confidence (0.85), cites precedent and policy friction. (Correct)\\[3pt]
\textbf{News model (b):} YES, high confidence (0.65), flips after reading late-April and May headlines suggesting tariffs were “likely” to rise to 150\%. (Wrong)
\end{quote}

Headlines indicated possibility rather than policy. The correct resolution required actual implementation by the deadline, which did not occur. Rumour anchoring overweighted momentum of coverage and underweighted institutional lag, shifting the model from a cautious, process-aware NO to an overconfident, headline-driven YES. Reasoning trace~(see Appendix~\ref{app:rumour}).

\textbf{Definition Drift.}
Models sometimes misinterpret acronyms or context when additional news shifts their semantic grounding, leading to incorrect predictions.

\begin{quote}
\textit{Question: “Will MATS applications open in March?”}\\[2pt]
\textbf{True resolution:} YES\\[3pt]
\textbf{Raw model (a):} YES, 0.58 confidence, interprets MATS as a recurring academic program that historically opens applications each March, referencing prior cycles. (Correct)\\[3pt]
\textbf{News model (b):} NO, 0.35 confidence, reinterprets MATS as the Mid-America Trucking Show, where registrations open months before March. (Wrong)
\end{quote}

The model with news context was exposed to snippets dominated by the trucking expo the most search visible meaning of MATS and thus shifted semantic grounding from an academic program to a trade event. This altered both the reference domain and the expected timeline, leading to a confident but misplaced “NO.” It underweighted contextual cues from the original question (application cycle, academic phrasing) and overtrusted frequency in retrieved snippets, effectively letting entity salience override contextual fit. Complete examples are provided~(see Appendix~\ref{app:defdrift}).

An additional behavior observed in \textit{DeepSeek-R1} \cite{deepseekr1} is that it does not provide any reasoning traces even when explicitly mentioned to verbalise its reasoning process through the prompt. The model does output the final tags in the form of \texttt{<answer>}\texttt{</answer>} and 
\texttt{<conf>}\texttt{</conf>} tags, but does not verbalise it's reasoning process. For examples~(see Appendix~\ref{app:deepseek})

\section{Conclusion}
Forecasting competence in LLMs is highly uneven, reflecting not only data coverage but the cognitive framing embedded in prompts. While we may expect adding recent news should improve forecasting accuracy, we find that sometimes it does while at other times it makes it worse because of definition drift, rumour anchoring and recency bias etc. The findings underscore that future-reasoning ability is conditional, not emergent, and invite design of benchmarks that disentangle knowledge recall from probabilistic inference.

\appendix
\section{Appendix}
\setcounter{secnumdepth}{2}
\renewcommand\thesection{\Alph{section}}
\renewcommand\thesubsection{\Alph{section}.\arabic{subsection}}

\section{Prompt Templates}\label{app:prompts}

\subsection{Core Forecasting Prompt (No News)}
\label{app:coreprompt}
\begin{quote}
\begingroup\ttfamily\footnotesize\sloppy
SYSTEM\_PROMPT = """\par
Question created on \{date\_str\}): \{question\}\par

Instructions:\par
1.\;Given the above question, rephrase and expand it to help you do better answering.\par
Maintain all information in the original question.\par
\textless rephrased\_question\textgreater \textless\\/rephrased\_question\textgreater\par

2.\;Provide a few reasons why the answer might be no. Rate the strength of each reason.\par
\textless no\_thoughts\textgreater\textless/\!no\_thoughts\textgreater\par

3.\;Provide a few reasons why the answer might be yes. Rate the strength of each reason.\par
\textless yes\_thoughts\textgreater\textless/\!yes\_thoughts\textgreater\par

4.\;Aggregate your considerations. Think like a superforecaster (e.g. Nate Silver).\par
\textless considerations\textgreater\textless/\!considerations\textgreater\par

5.\;Output an initial probability (prediction) given steps 1--4.\par
\textless initial\_probability\textgreater\textless/
\/ initial\_probability\textgreater\par

6.\;Evaluate whether your calculated probability is excessively confident or not confident enough. Also,\par
consider anything else that might affect the forecast that you did not before consider.\par
\textless extra\_considerations\textgreater\textless/
\/extra\_considerations\textgreater\par

7.\;Output your answer in \textless ans\textgreater YES/NO \textless/\!ans\textgreater\;and the confidence in \textless conf\textgreater\;0--1\; \textless/\!conf\textgreater.\par
Output the confidence as a number between 0 and 1 (e.g. 0.85), without a \% sign. Do not output anything else.\par
Make sure you follow all instructions, reduce reasoning effort if required. Do not repeat the points mentioned in the prompt.\par

Example (correct format):\par
\textless extra\_considerations\textgreater\par
While I am reasonably confident in this forecast, unexpected events such as corporate restructuring,\par
internal conflicts, or external pressures could alter the situation. The absence of recent news increases\par
confidence, but I remain cautious due to the unpredictable nature of corporate leadership dynamics.\par
\textless/\!extra\_considerations\textgreater\par

Example (wrong format):\par
6.\;Evaluate whether your calculated probability is excessively confident or not confident enough. Also,\par
consider anything else that might affect the forecast that you did not before consider.\par
\textless extra\_considerations\textgreater\par
While 0.85 confidence is high, it is not overly confident because unexpected events can always intervene.\par
However, no recent news or rumors indicate instability in leadership. The media industry can be volatile,\par
but major leadership changes often come with early signals. Since none are evident, the estimate seems\par
reasonable. It would be prudent to slightly discount the confidence if any new information arises during 2024,\par
but as of now, 0.85 is appropriate.\par
\textless/\!extra\_considerations\textgreater\par
"""\par
\endgroup
\end{quote}

\subsection{News-Augmented Forecasting Prompt}
\label{app:newsprompt}
The forecasting prompt used in this condition is identical to the base prompt described in Appendix~A.1.  
The only modification is that up to ten time-bounded news snippets retrieved via the \textit{Exa} API~\cite{exa2025api} 
are appended to the end of the prompt before model inference.
\subsection{Category Classification Judge Prompt}
\label{app:classifier}
\begin{quote}
\begingroup\ttfamily\footnotesize\sloppy
SYSTEM\_PROMPT = """\par
\# Question Classifier System Prompt\par

You are a classifier that categorizes a given question into one of the following categories and their respective sub-categories. Choose the most appropriate category and sub-category that best fits the question's primary focus.\par

\#\# Categories and Sub-Categories:\par

\#\#\# 1. Politics\par
- Domestic Policy: Questions about internal government policies, legislation, regulations, and governance within a country.\par
- Elections \& Campaigns: Questions about voting processes, political candidates, election results, and campaign activities.\par
- Political Parties \& Ideologies: Questions about political movements, party platforms, political philosophies, and partisan issues.\par
- Government Structure: Questions about constitutional matters, branches of government, political systems, and institutional processes.\par
- Public Policy \& Social Issues: Questions about policy debates, social reforms, civil rights, and politically relevant social topics.\par

\#\#\# 2. Entertainment\par
- Movies \& Television: Questions about films, TV shows, streaming content, actors, directors, and cinema industry.\par
- Music \& Audio: Questions about songs, artists, albums, concerts, music industry, and audio entertainment.\par
- Gaming: Questions about video games, gaming platforms, esports, game development, and gaming culture.\par
- Celebrity \& Pop Culture: Questions about famous personalities, entertainment news, awards, and popular culture trends.\par
- Books \& Literature: Questions about authors, novels, publishing, literary works, and reading culture.\par

\#\#\# 3. Sports\par
- Professional Sports: Questions about major league competitions, professional athletes, team performance, and sports business.\par
- International Competitions: Questions about Olympics, World Cup, continental championships, and global sporting events.\par
- Individual Sports: Questions about tennis, golf, athletics, martial arts, and other individual competitive activities.\par
- Team Sports: Questions about football, basketball, cricket, rugby, and other team-based sports.\par
- Sports Culture \& Recreation: Questions about sports history, fan culture, recreational activities, and sports lifestyle.\par

\#\#\# 4. Technology\par
- Computing \& Software: Questions about computers, operating systems, applications, programming, and software development.\par
- Internet \& Digital Services: Questions about websites, online platforms, digital services, and internet-related topics.\par
- Mobile \& Consumer Electronics: Questions about smartphones, tablets, gadgets, and consumer technology products.\par
- Emerging Technologies: Questions about artificial intelligence, blockchain, virtual reality, and cutting-edge innovations.\par
- Tech Industry \& Business: Questions about technology companies, tech entrepreneurship, and technology market trends.\par

\#\#\# 5. Finance\par
- Personal Finance: Questions about budgeting, saving, investing, loans, and individual financial planning.\par
- Banking \& Financial Services: Questions about banks, credit, insurance, financial institutions, and financial products.\par
- Markets \& Trading: Questions about stock markets, bonds, commodities, trading strategies, and investment vehicles.\par
- Economic Indicators: Questions about inflation, GDP, unemployment, interest rates, and macroeconomic metrics.\par
- Corporate Finance: Questions about business finance, company valuations, mergers \& acquisitions, and corporate financial strategies.\par

\#\#\# 6. Geopolitics\par
- International Relations: Questions about diplomatic relations, treaties, international cooperation, and bilateral/multilateral agreements.\par
- Global Conflicts: Questions about wars, territorial disputes, military tensions, and international security issues.\par
- Trade \& Economics: Questions about international trade, economic sanctions, trade agreements, and global economic relations.\par
- Regional Affairs: Questions about specific geographic regions, regional organizations, and area-specific political developments.\par
- Global Governance: Questions about international organizations, global institutions, international law, and worldwide policy coordination.\par

7.\;If it is not related to any of the above, classify it as ``Irrelevant'' for both category and sub-category.\par

\#\# Classification Instructions:\par
1.\;Read the question carefully and identify its primary focus.\par
2.\;Select the most appropriate main category (1--6).\par
3.\;Choose the most relevant sub-category within that main category.\par
4.\;If a question spans multiple categories, choose the one that represents the primary or most significant aspect.\par
5.\;STRICTLY STICK TO THE CATEGORIES MENTIONED. DO NOT MAKE UP ANY CATEGORY.\par
For the category field your answer should be \texttt{SPORTS, ENTERTAINMENT, POLITICS, GEOPOLITICS, TECHNOLOGY, FINANCE,} or \texttt{IRRELEVANT.} Nothing else.\par
"""
\endgroup
\end{quote}

\subsection{Refined Filtering Judge Prompt}
\label{app:filter}
The following is the detailed prompt for stage three of our data processing pipeline.
\begin{quote}
\begingroup\ttfamily\footnotesize\sloppy
"""\textbackslash\par
You are a strict *relevance* classifier for prediction-market questions.\par

Relevant (True):\par
- About a public, verifiable event/outcome/metric (elections, matches, earnings, court rulings, awards, macro prints, product releases, etc.).\par
Irrelevant (False):\par
- Personal/self-referential (e.g., ``Will I meet X…'', ``Will I score top in Kaggle…''), unverifiable private matters.\par
- Platform/meta mechanics (e.g., ``Will all 'YES' traders have a unique max payout…'').\par
- Vague opinion polls with no verifiable source.\par

Edge:\par
- Public figures/events/results/releases are Relevant if objectively checkable.\par

Return JSON only: \{"is\_relevant": true|false, "reason": "<one sentence>"\}.\par
"""
\endgroup
\end{quote}

\section{Failure Mode Case Studies}
\label{app:failmodes}
\subsection{Recency Bias Example: ``S\&P 500 above \$6050 on June 13?''}
\label{app:recency}
\begin{quote}
\begingroup\ttfamily\footnotesize\sloppy
\{"question": "S\&P 500 above \$6050 on June 13?", 
"true\_resolution": "NO", 
"pred\_1": "NO", 
"pred\_2": "YES", 
"correct\_1": true, 
"correct\_2": false, 
"conf\_1": 0.66, 
"conf\_2": 0.54, 
"response\_1": "<rephrased\_question>Will the S\&P 500 index be above 6,050 on June 13, 2025 (interpreting 'above' as the closing level on that date), as asked on June 11, 2025?</rephrased\_question><no\_thoughts>- Likely starting level below 6,050; reaching/exceeding it within two trading days requires an above-average move. Strength: High - 6,000 is a psychological/resistance area; initial breakouts often fail or retest before advancing. Strength: Medium - Near-term macro/event risk could trigger a pullback (e.g., data releases, Fed communications). Strength: Medium - Mean reversion after recent gains can cap short-horizon upside. Strength: Medium-Low - Valuation headwinds reduce odds of rapid further gains over a 2-day window. Strength: Low-Medium</no\_thoughts><yes\_thoughts>- Ongoing bull-market momentum, especially in mega-cap tech, can push the index through round-number levels. Strength: Medium - A clean breakout above 6,000 could trigger CTA/stop-buying and extension to 6,050+. Strength: Medium - Positive macro surprises (easing inflation, supportive policy) can add 1\%+ in a day. Strength: Low-Medium - Low-volatility grind-up regimes increase chances of incremental new highs. Strength: Low-Medium</yes\_thoughts>
 <considerations>Without the exact June 11 level, scenario analysis: - If ~5,800: odds to be >6,050 by June 13 are low (~10\%). - If ~5,900–5,980: moderate (~25–40\%), needing a 1--2\% move. - If already greater than equal to 6,000: moderate-to-good (~50–65\%), as momentum can extend. Plausible mid-2025 baseline places the index more likely below 6,000 than above, so weighted odds favor “No.” Aggregating scenarios yields an estimate near one-third for “Yes.”</considerations>
 <initial\_probability>
 0.34</initial\_probability>
  <extra\_considerations>Uncertainty about whether 'above' refers to close versus intraday; using close lowers the probability slightly. Lack of the exact starting level introduces major variance; to avoid overconfidence, keep the estimate near one-third rather than lower. No further strong asymmetries identified.</extra\_considerations>
   <ans>NO</ans><conf>0.66</conf>", 
"response\_2": "<rephrased\_question>Will the S\&P 500 index finish above 6,050 on Friday, June 13, 2025? Assume 'above 6,050' refers to the official closing level of the S\&P 500. Context: the index recently revisited ~6,000 (first time since February) and several strategists (e.g., Barclays) raised bullish year-end targets to 6,050.</rephrased\_question><no\_thoughts>- Event risk (CPI/Fed around June 12 could be hawkish, push yields up, hit equities): Strong - 6,000–6,050 is a psychological/technical resistance zone inviting profit-taking/pinning: Medium-Strong - Recent run-up increases odds of mean reversion/‘sell the news’: Medium - Narrow leadership/valuation stretch raises fragility to negative surprises: Medium-Weak</no\_thoughts><yes\_thoughts>- Positive macro surprise (benign CPI/dovish Fed tone) can produce >1\% rally, clearing 6,050: Strong - Momentum/breakout dynamics after reclaiming 6,000 with bullish sentiment and upgrades: Medium - Mega-cap tech leadership can drive index-level gains quickly: Medium - Trend-following/systematic flows post-upgrade headlines support near-term upside: Weak-Medium</yes\_thoughts>
 <considerations>Base rate: with the index near ~6,000, a 0.8\% move to >6,050 over two sessions is common, ~45–55\% territory. Macro catalysts create a bimodal outcome; if CPI/Fed are market-friendly, a decisive break above 6,050 is plausible; if not, index may stall or retrace to/below 6,000 due to resistance and positioning. Sentiment skew is modestly positive given recent upgrades, but event risk tempers confidence. Slight edge to “Yes.”</considerations>
  <initial\_probability>0.55
   </initial\_probability>
   <extra\_considerations>Ambiguity between closing level vs. intraday high matters; intraday probability would be higher (~0.60–0.65). Assuming close, options pinning around 6,000 could cap gains. Event clustering increases variance, so 0.55 may be a touch high; adjust slightly down to 0.54 to account for pinning/mean reversion risk.</extra\_considerations>
    <ans>YES</ans><conf>0.54</conf>"\}
\endgroup
\end{quote}

\subsection{Rumour Overweighting: ``Tariffs on China above 150\% by end of June?''}
\label{app:rumour}
\begin{quote}
\begingroup\ttfamily\footnotesize\sloppy
\{"question": "Tariffs on China above 150\% by end of June", 
"true\_resolution": "NO", 
"pred\_1": "NO", 
"pred\_2": "NO", 
"correct\_1": true, 
"correct\_2": true, 
"conf\_1": 0.88, 
"conf\_2": 0.58, 
"response\_1": "<rephrased\_question> Will any government impose tariffs on imports from China at a rate exceeding 150\% by the end of June (June 30, 2025)? Ambiguity notes: “Tariffs on China above 150\%” could mean (a) broad headline tariffs by a major economy (e.g., the U.S.) on a major product category (such as EVs), (b) average tariffs across goods, or (c) product-specific duties (including anti-dumping/countervailing duties) that sometimes exceed 150\%. The timeframe is short (from April 11 to June 30, 2025).</rephrased\_question><no\_thoughts>- Policy/process friction: large headline tariff hikes (e.g., Section 301/232-type moves) typically require procedures, consultations, and political capital; implementing greater than 150\% in close to 2.5 months is unlikely. Strength: High - Political-economic costs: a 150\%+ tariff invites retaliation, supply-chain disruption, and inflationary risks; leaders may prefer targeted bans/controls instead. Strength: High - Existing measures already high: where protection is desired (e.g., EVs), tariffs were already raised substantially; further quick escalation to greater than \!150\% risks overreach. Strength: Medium-High - Alternatives to tariffs: governments can use export controls, investment restrictions, or outright bans (national security) rather than extreme tariff rates. Strength: Medium - International/legal constraints: WTO norms and allied coordination temper extreme moves absent acute crisis. Strength: Medium</no\_thoughts><yes\_thoughts>- Product-specific trade remedies: anti-dumping/countervailing duty cases can yield provisional rates well above 150\% on specific Chinese products. Strength: Medium - Geopolitical shock: a sharp escalation (e.g., security incident) could trigger emergency national-security tariffs or bans with very high rates. Strength: Low-Medium - Domestic industry pressure: politically salient sectors (autos, steel, solar) could spur rapid action, especially if framed as national security. Strength: Low-Medium - Legislative action: a fast-tracked bill could mandate punitive tariffs, though timing is tight. Strength: Low - Precedent for firm-specific rates: certain firms can face >\!150\% AD/CVD margins; if the question counts these as “tariffs on China,” this could satisfy the condition. Strength: Medium</yes\_thoughts><considerations>Key uncertainty is definition. If counting any product-specific AD/CVD, base rates for some Chinese goods exceeding 150\% exist and new cases could cross 150\% within months. If the intent is a broad, headline tariff (e.g., on EVs) exceeding 150\%, the base rate is extremely low in peacetime, and the short horizon lowers odds further. Political incentives favor targeted tools or security-based restrictions over shock 150\%+ tariff hikes. Absent a major geopolitical trigger or imminent trade case known to produce such margins, the most likely outcome by end-June is no new 150\%+ headline tariff; niche AD/CVD outcomes are possible but uncertain within the timeframe.</considerations>
  <initial\_probability>0.12
   </initial\_probability>
   <extra\_considerations>Main risk is definitional: if the market resolves based on any AD/CVD rate exceeding 150\% on any Chinese product, probability should be higher. If it requires a broad, high-profile tariff (e.g., EVs), 0.12 may be generous. The short window further reduces odds. My estimate leans conservative; uncertainty remains about ongoing trade cases not captured here, so confidence in NO is moderate-high, not absolute.</extra\_considerations>
     <ans>NO</ans><conf>0.88</conf>", 
"response\_2": "<rephrased\_question>By June 30, 2025, will the United States’ headline tariff rate applied to merchandise imports from China be increased to greater than 150\% (i.e., at least 151\%)? Current context suggests recent hikes variously reported as  approximately 54\%, 125\% “reciprocal,” and up to 145\% per some sources, including changes targeting de minimis shipments (e.g., Temu/Shein). Assess whether a further hike above 150\% will be enacted by end-June.</rephrased\_question>
 <no\_thoughts>- Political/economic blowback: additional hikes above 150\% could worsen inflation and consumer prices; strong business and retail opposition likely to slow moves. Strength: High - Legal/administrative frictions: further increases may require additional proclamations, inter-agency review, and risk injunctions or litigation; timing to end-June is tight. Strength: Medium - Recent large increases suggest a pause: after big April moves (to 125–145\%), the administration may assess impacts before another escalation. Strength: Medium - Diplomatic/strategic pacing: keeping headroom for future leverage rather than immediate escalation beyond 150\%. Strength: Low-Medium</no\_thoughts><yes\_thoughts>- Policy posture: the administration’s hawkish stance and willingness to move rapidly in April imply readiness for further hikes soon. Strength: Medium-High - Executive authority: tools like Section 301/IEEPA/“reciprocal tariffs” allow swift adjustments without Congress, enabling a quick bump above 150\%. Strength: Medium - Targeting de minimis/fast-fashion imports: further tightening to curb Temu/Shein could include pushing rates above 150\%. Strength: Low-Medium - Signaling/politics: a symbolic breach above 150\% could be touted as toughness on China. Strength: Low</yes\_thoughts>
  <considerations>Conflicting reports (54\%, 125\%, 145\%) imply definitional ambiguity; assuming the “headline” broad tariff. Moving from close to 145\% to greater than \!150\% within 80 days would be another significant step with inflation optics and legal risks. While the White House can act quickly, successive hikes risk backlash and court challenges. The administration may prefer enforcement and implementation of April measures before escalating again. A move exactly to 150\% (not greater than \!150\%) is more politically “clean,” but the question requires greater than 150\%, which is less likely unless jumping to a higher round (e.g., 200\%), which seems aggressive by June.</considerations>
   <initial\_probability>0.42
    </initial\_probability>
     <extra\_considerations>Ambiguity over which tariff measure is referenced could sway the outcome: certain product-specific or penalty rates may already exceed 150\%, but the market narrative tends to cite a single headline rate. Court challenges or preliminary injunctions could delay any new proclamation. If economic data in May–June show easing inflation, political space for escalation could widen; conversely, any price spikes or supply-chain disruptions could constrain action. Given these, 0.42 may be slightly conservative but avoids overconfidence amid definitional and legal uncertainties.</extra\_considerations>
      <ans>NO</ans>
       <conf>0.58</conf>"\}
\endgroup
\end{quote}

\subsection{Definition Drift: ``Will MATS applications open in March?''}
\label{app:defdrift}
\begin{quote}
\begingroup\ttfamily\footnotesize\sloppy
\{"question": "Will MATS applications open in March?", 
"true\_resolution": "YES", 
"pred\_1": "YES", 
"pred\_2": "NO", 
"correct\_1": true, 
"correct\_2": false, 
"conf\_1": 0.58, 
"conf\_2": 0.35, 
"response\_1": "<rephrased\_question>Are the upcoming MATS program applications expected to open in March (as in prior cycles), or has the application timeline shifted?</rephrased\_question> 
<no\_thoughts> - The program could shift its calendar due to funding or operational changes; March might move to Feb/Apr (strength: medium). - They might adopt rolling admissions instead of a fixed March opening (strength: medium). - Ambiguity about which “MATS” is meant; some programs with that acronym do not open in March (strength: strong). - External constraints (venues, visas, partner timelines) could push the window later (strength: weak–medium). 
</no\_thoughts> 
<yes\_thoughts> - Historical pattern: recent MATS summer cohorts have typically opened applications around March (strength: strong). - March aligns well with a summer-start pipeline, making it a stable choice (strength: medium). - Absent contrary announcements, organizations tend to keep predictable cycles (strength: medium). 
</yes\_thoughts> 
<considerations>Base rates from prior cycles favor March for the summer round, but acronym/program ambiguity and potential operational changes reduce confidence. With no specific year given and no fresh signals, a modestly favorable probability is warranted.</considerations> 
<initial\_probability>0.6
 </initial\_probability> 
<extra\_considerations>Ambiguity likely makes the estimate slightly overconfident. Nudging down to reflect possible reference to a different “MATS” or a one-off schedule change. Final probability: 0.58.</extra\_considerations> 
<ans>YES</ans> 
<conf>0.58</conf>", 
"response\_2": "<rephrased\_question>
 Clarify whether the “MATS” in question refers to the Mid-America Trucking Show and, if so, whether its application/registration window (for exhibitors or attendees) is scheduled to open in March of the relevant year. If “MATS” refers to a different program with the same acronym, is its application window expected to open in March?</rephrased\_question> 
<no\_thoughts> - The Mid-America Trucking Show (MATS) event typically occurs in late March, so exhibitor applications and attendee registration usually open months earlier (Strong). - A recent 2025 reference shows MATS activities happening around late March/early April, implying applications would have opened before March (Moderate-Strong). - “Applications” may not be the right term for MATS attendee registration; exhibitor applications and media credentials typically open well in advance (Moderate). - If the user meant AIMA MAT (not MATS), those application cycles for February/March sessions tend to open in January/February, not start in March (Weak-Moderate due to acronym mismatch). 
</no\_thoughts> 
<yes\_thoughts> - Some organizations open final or late application windows in March for events occurring late March/early April; MATS could allow late registrations or on-site sign-ups (Weak-Moderate). - If “MATS” refers to a different program with that acronym, its cycle could plausibly begin in March (Weak). - Exceptional scheduling changes (e.g., policy shifts, one-off timelines) could push openings into March in a given year (Weak). 
</yes\_thoughts> 
<considerations>Base rate: large trade shows like MATS open exhibitor/attendee processes months ahead; March openings are atypical because the event is in March. Evidence: 2025 reporting places MATS activity around late March/early April; consistent with prior years, reducing likelihood of March opening. Ambiguity: acronym uncertainty increases variance, but the most common “MATS” is the trucking show. Tail scenarios: late/last-minute windows exist but are not the primary “applications opening” phase.</considerations> 

<initial\_probability>0.30
 </initial\_probability> 
<extra\_considerations>Ambiguity could mean the user is asking about a different MATS entirely, which modestly raises the chance of a March opening. If referring specifically to on-site registration or late credentialing, “opening in March” could be technically true, though not the main phase. To avoid overconfidence in an ambiguous acronym, slight upward adjustment from 0.30 to 0.35 for YES seems appropriate.</extra\_considerations> 
<ans>NO</ans> 
<conf>0.65</conf>"\}
\endgroup
\end{quote}

\subsection{Missing Reasoning Traces in DeepSeek-R1}
\label{app:deepseek}
\begin{quote}
\begingroup\ttfamily\footnotesize\sloppy
\{"question": "NFL Week 17 (Dec 28 2024): Will the Washington Commanders (home) beat the Atlanta Falcons (away)?", 
"true\_resolution": "YES", 
"pred\_1": "YES", 
"pred\_2": "YES", 
"correct\_1": true, 
"correct\_2": true, 
"conf\_1": 0.7, 
"conf\_2": 0.7, 
"response\_1": "<ans>YES</ans> <conf>0.7</conf>", 
"response\_2": "<ans>YES</ans> <conf>0.7</conf>"\}
\endgroup
\end{quote}

\section*{Acknowledgments}
We thank Lossfunk \cite{lossfunk2025} for its support throughout this project. This work benefited from access to compute resources provided by Lossfunk, as well as constructive discussions and idea-level guidance that helped shape the direction of this research.

\bibliography{aaai2026}

\end{document}